\documentclass[letterpaper, 10 pt, conference]{ieeeconf}  

\IEEEoverridecommandlockouts                              

\usepackage{graphicx}

\title{\LARGE \bf
Saliency-Based Attention Shifting: A Framework for Improving Driver Situational Awareness of Out-of-Label Hazards
}

\author{Yousra Shleibik$^{*}$, Jordan Sinclair$^{*}$ and Kerstin Haring$^{*}$
\thanks{$^{*}$ The authors are with the Ritchie School of Engineering and Computer Science, University of Denver, Denver, CO, USA.}%
\thanks{
        {}}%
}

\begin{document}

\maketitle
\thispagestyle{empty}
\pagestyle{empty}

\begin{abstract}

The advent of autonomous driving systems promises to transform transportation by enhancing safety, efficiency, and comfort. As these technologies evolve toward higher levels of autonomy, the need for integrated systems that seamlessly support human involvement in decision-making becomes increasingly critical. Certain scenarios necessitate human involvement, including those where the vehicle is unable to identify an object or element in the scene, and as such cannot take independent action. Therefore, situational awareness is essential to mitigate potential risks during a takeover, where a driver must assume control and autonomy from the vehicle. The need for driver attention is important to avoid collisions with external agents and ensure a smooth transition during takeover operations. This paper explores the integration of attention redirection techniques, such as gaze manipulation through targeted visual and auditory cues, to help drivers maintain focus on emerging hazards and reduce target fixation in semi-autonomous driving scenarios. We propose a conceptual framework that combines real-time gaze tracking, context-aware saliency analysis, and synchronized visual and auditory alerts to enhance situational awareness, proactively address potential hazards, and foster effective collaboration between humans and autonomous systems.

\end{abstract}

\section{INTRODUCTION}

As autonomous driving technology improves, reaching higher degrees of independence, the need for integrated technologies that facilitate more seamless interactions to support human participation in decision making continues to grow. In addition to recognizing when a takeover request (TOR) should be initiated, self-driving vehicle technologies need to provide a streamlined method for the driver to gain situational awareness and make informed decisions when faced with unlabeled obstacles. This highlights the critical role of human-vehicle interaction (HVI) in bridging the gap between current capabilities and the goal of synchronous teaming between an autonomous vehicle and a driver. This is challenging to achieve with outside factors like driver distractions or non-driving related tasks (NDRTs) that take attention away from the road and force additional time to be spent understanding the context behind a TOR. 

\begin{figure}[h!]
    \centering
    \includegraphics[width=1.0\linewidth]{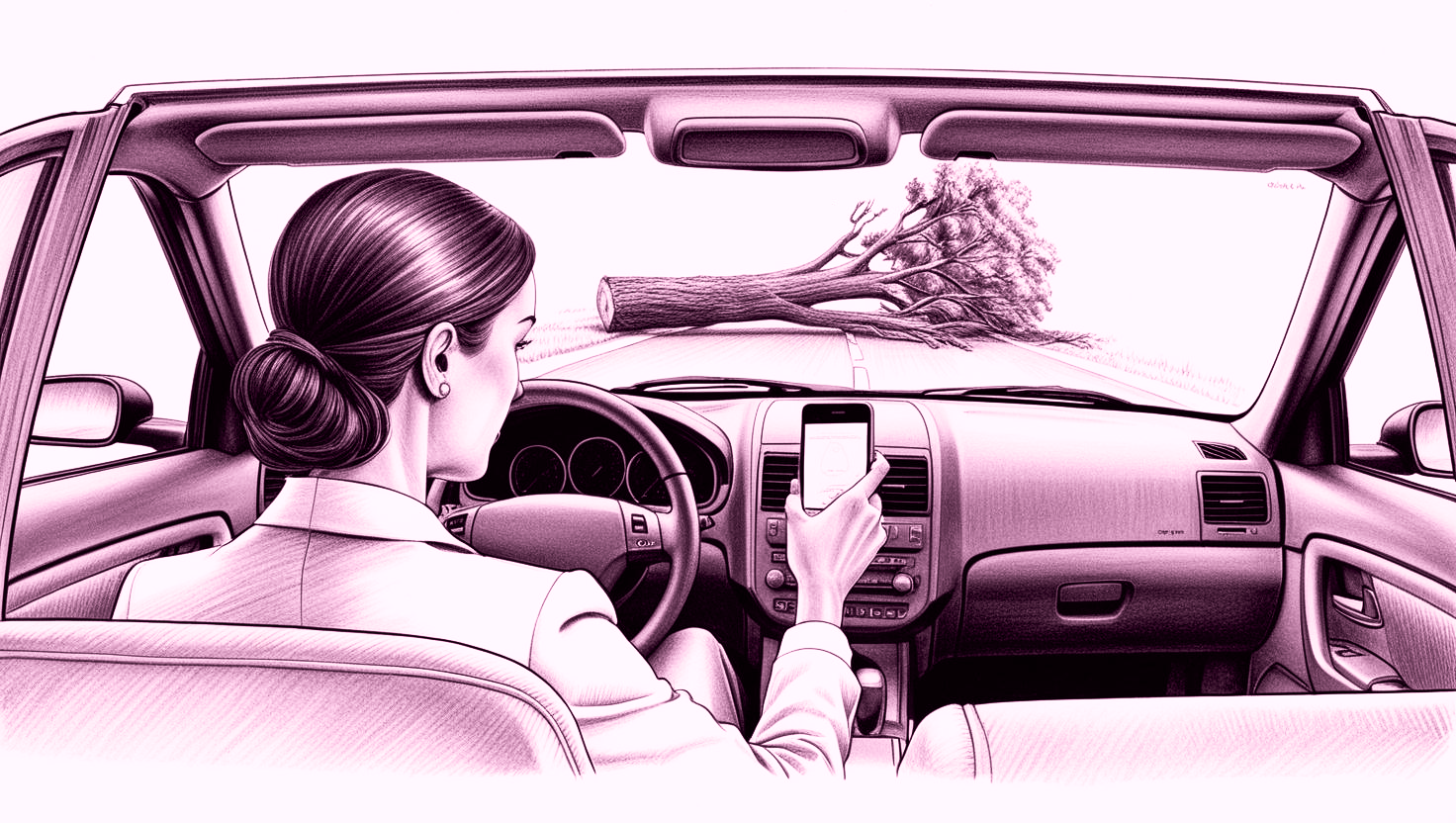}
    \caption{Illustration of the problem our approach tackles. The driver of an autonomous vehicle is obviously distracted, and there is an unlabeled hazard in which the vehicle requests the driver's control. 
    }
    \label{fig:scenario}
    \vspace{-8pt}
\end{figure}

Both the vehicle and the driver must act with agency in a self-driving system, and thus both must share knowledge of road events. The driver must exert influence over the vehicle at times in order to provide instructions for appropriate action, and the vehicle also guides the driver through its automated systems. This distributed knowledge and task sharing structure is often considered similar to the metaphor of a horse and rider \cite{distributed}. 
However, the driver cannot maintain the tactical intelligence necessary to successfully take control of the vehicle when distracted by NDRTs, such as reading or using a phone or computer \cite{autoexp}. Despite posing several dangers, this type of behavior is probable to occur in "self-driving" cars, as users will feel comfortable performing unrelated activities if they do not experience a negative TOR. Infrastructure must be considered to better handle TORs under different in-cabin scenarios towards the ability for an automated system to move past unlabeled hazards. 

We seek to address target fixation and driver readiness in autonomous vehicle takeovers by answering the following questions.
\begin{enumerate}
    \item What is the initial position of the driver's visual focus when distracted from the driving task?
    \item What environmental elements are essential for restoring the driver's situational awareness?
    \item How can the driver's attention be directed from their distraction point to the hazard location?
\end{enumerate}
We propose a system that relies on gaze information from the driver to understand where they are looking and need to be looking for a fast response. Based on this, visual and audio cues are presented to the driver to break their focus on outside tasks and to quickly indicate the object or event identified by the automated vehicle. Ultimately, the driver's attention should be redirected to the most critical aspects of the scene, reducing the time it takes to regain situational awareness during a TOR.

\section{Background \& Problem Formulation}

Self-driving vehicles operate at varying levels of autonomy, each requiring a different degree of human oversight \cite{alshami2024smart}. Most current autonomous vehicles operate at SAE Level 3, where the system handles most driving tasks but requires human intervention in unexpected situations, such as encounters with out-of-label objects \cite{epa_self_driving_vehicles}. The critical and spontaneous cooperation required to address such road events can be challenging to obtain without well-defined communication interfaces, such as through visual, auditory, and kinesthetic cues from both the driver and the vehicle. Without these signals, driver takeovers lead to fatigue, inaccuracy, and latency in reaction times \cite{physiological}. In addition, the act of transferring power between an SDC and the driver often causes scattered gaze or fixation on certain environmental features that may be unrelated to the task \cite{MERAT2014274}.

Target fixation causes an individual to become hyper-focused on one object or region, often incapable of tackling a new task without some form of disruption. 
This worsens the impact of takeovers from an automated vehicle, which can already be 30-40 seconds without systems to ease the transition to autonomy and control \cite{MERAT2014274}. The human driver could have a very limited time to gain situational awareness of the driving environment. That stress and pressure, particularly if unprecedented through an understandable guidance system, can lead to driver fixation on the most salient element in their field of view (FoV), such as an unrecognized object or pedestrian, for example. This narrow focus can reduce situational awareness and the ability to choose safe evasive maneuvers. For example, instead of considering alternative lanes, escape routes, or adjusting vehicle speed, a driver who has 'locked onto' a target can inadvertently steer toward it or do not notice secondary hazards. 
Studies \cite{wang2022dual, biswas2024modeling} in driver attention and eye tracking have shown that the distribution of the driver's gaze is essential to maintaining robust situational awareness. Thus, the intensity of target fixation can be inversely related to preparedness for safe maneuvering in emergency scenarios.


The presence of target fixation and other physiological impacts on the driver suggests the need for improved takeover systems with clearer demonstrations of information. Human-Machine Interface (HMI) design can play a role in the mitigation of such issues in a variety of ways, including: 

\begin{itemize}
    \item \textbf{Multimodal Alerts:} Using audible and tactile cues, rather than purely visual alerts, can encourage drivers to break their fixation and broaden their attention \cite{einhauser2024high}.

    \item \textbf{Gaze Monitory \& Guidance:} Some research is exploring how to infer driver attention \cite{wang2022dual}. There is also research that seeks to subtly guide or redirect gaze to critical areas \cite{shleibik2024towards}. This could be applied to highlight key features in a vehicle, such as mirrors or peripheral zones, through smart displays, light bars, or augmented reality cues.

    \item \textbf{Adaptive Warning Systems:} Systems that monitor driver gaze behavior can detect prolonged fixation on a single point \cite{wang2022dual}. If the system notices that the driver is not scanning effectively, it could prompt scanning behavior or reorient the driver’s attention.
\end{itemize}



\begin{figure}[tb]
    \centering
    \includegraphics[width=1.0\linewidth]{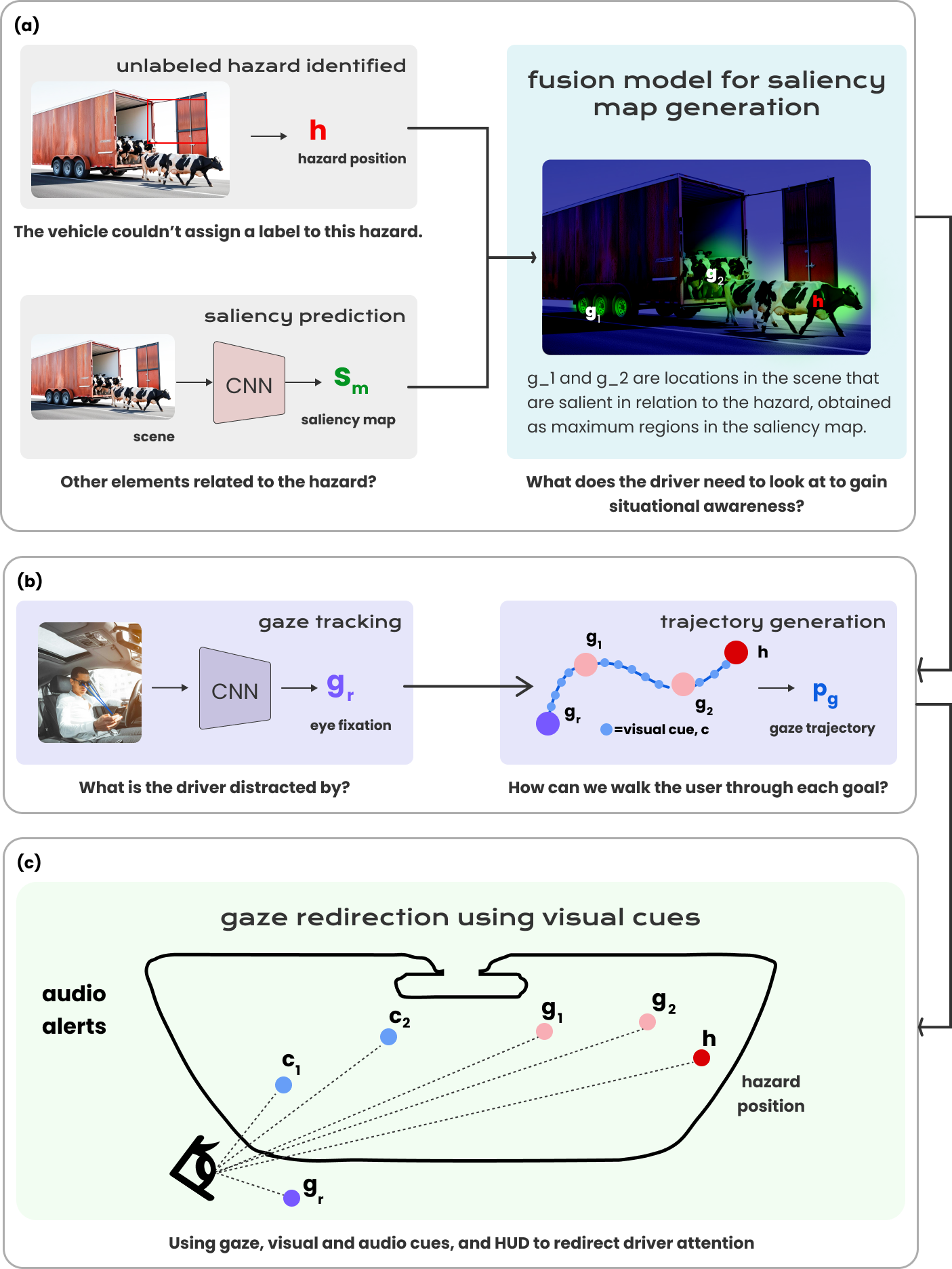}
    \caption{This system diagram indicates what will happen after an unlabeled hazard is encountered. (a) illustrates the procedure of obtaining a filtered saliency map based on the hazard location. (b) describes how a trajectory is generated between the driver's initial gaze and the hazard location, including any highly salient regions. (c) displays the trajectory to the driver using an HUD.}     
    \label{fig:flowchart}
    \vspace{-8pt}
\end{figure}

\section{APPROACH}

We propose a conceptual framework that integrates real-time gaze tracking, context-aware saliency analysis, and coordinated visual and auditory cues to enhance driver attention during scenarios with unlabeled hazards. By bringing these components together into a single integrated method, we ensure that the driver gains situational awareness about the most pressing aspects of the environment during an unexpected takeover request. The elements of our approach are detailed in the following sections, including the techniques used to estimate the saliency map, track gaze, the design and integration of HUD visual cues, and the role of audio alerts in strengthening driver situational awareness. On a high level, the system recognizes the need of a take-over, the target fixation of the human, and interrupts the fixation by redirecting attention to the more relevant area. The integration of these elements fosters a more responsive human-vehicle interaction, expected to ultimately advance road safety. Figure \ref{fig:flowchart} illustrates the proposed system.

\subsection{Saliency Map Generation}


Saliency maps are utilized to increase the human-likeness of autonomous driving by focusing attention on the most relevant areas in a scene. However, they are inefficient without additional information, like driver gaze or depth \cite{looking}. Similarly, Hu, et al. \cite{hu2024context} introduced a fusion model that transforms an initial saliency map using driver gaze information to target only the most driver-relevant components of the scene. We leverage this fusion model as the backbone of our system, where the input is instead the position of an unlabeled hazard identified by an assumed anomaly detection algorithm \cite{10419101}. From this modified saliency mapping, we take high-value regions mapped to object locations $g_i$. These regions have high importance in relation to the hazard $h$, providing information about potential upcoming events or ramifications of driver action. In order to gain situational awareness prior to the takeover of an autonomous vehicle, the driver should pay attention to the highlighted regions of this saliency map. 

\subsection{Gaze Tracking \& Attention Shifting }
\label{section:Gaze}

Accurately determining real-time driver attention begins with an inward-facing camera that continuously captures facial features and pupil position.  Advanced eye-tracking algorithms then estimate the gaze vector and fixation points, providing a detailed record of where the driver is looking at any given moment. We use this information to guide the driver's gaze to points of interest for a given takeover scenario, similar to the use of visual guidance discussed by Shleibik, et al. \cite{shleibik2024towards}. This paper explored the shift of attention and the attraction of attention in a robotic interaction scenario, leveraging an AR headset and a ground robot to improve situational awareness in dynamic environments. Using gaze tracking, the system seamlessly shifts or attracts the user’s attention to relevant points of interest using markers, which are updated when the user's gaze lands on each marker. In our approach, we employ a similar method to Shleibik, et al. \cite{shleibik2024towards}, in using visual cues positioned in the FoV of the driver to continuously redirect their attention towards the hazard. We generate a trajectory from the driver's current fixation point to the hazard location while optimizing additional information relevant to decision-making during a takeover. Thus, we shift the driver's gaze to multiple critical areas related to the initial hazard. 
In Fig. \ref{fig:flowchart}, we demonstrate an example of our method where two potential hazards, $g_1$ and $g_2$, are identified from the fused saliency map created using unlabeled hazard, $h$, that could change how the driver should react when taking over the vehicle. These potential hazards are integrated as waypoints when generating a trajectory of visual cues (see Fig. \ref{fig:flowchart}-b) to guide the driver's gaze to the hazard, thus improving situational awareness. We utilize head-up display (HUD) technology to present these visual cues, discussed in further detail in Section \ref{section:HUD}.




\subsection{Head-up Display (HUD) \&  Visual Cues}
\label{section:HUD}

\begin{figure}
    \centering
    \includegraphics[width=0.9\linewidth]{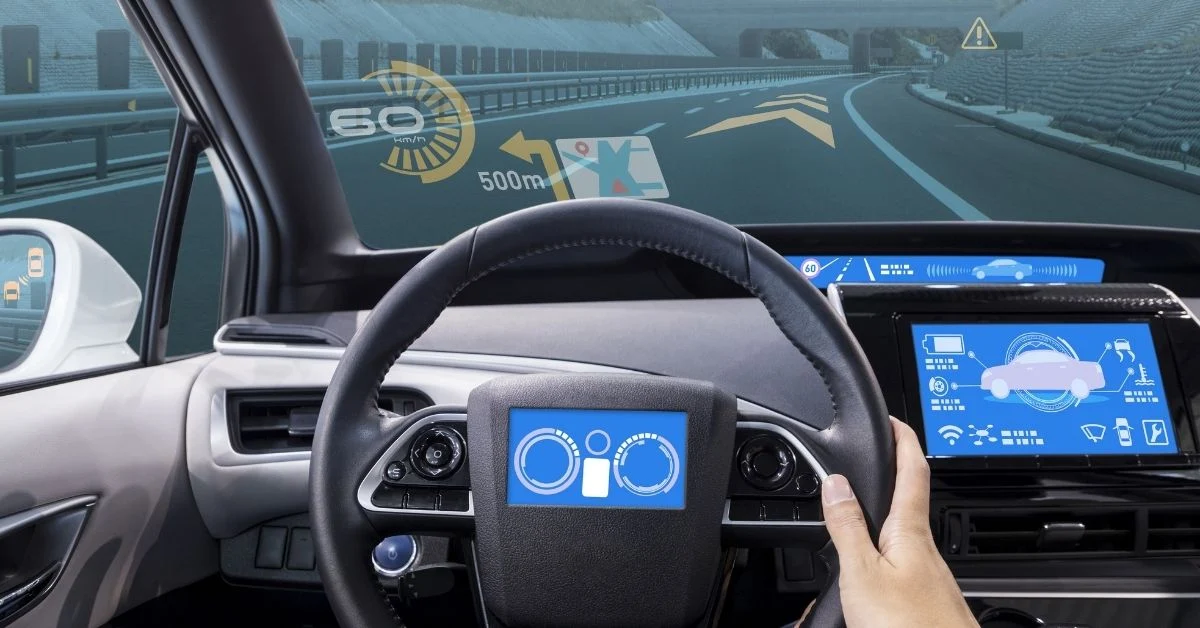}
    \caption{Head-Up Display (HUD) Integration, source: \cite{hud}}
    \label{fig:HUD}
    \vspace{-8pt}
\end{figure}

Head-Up Displays (HUDs) cast vital information onto a vehicle's windshield, allowing a driver to stay focused on the road. In our conceptual system, the HUD functions as an integral interface for delivering real-time visual guidance to help drivers gain situational awareness and respond swiftly to potential hazards. 
By projecting sequential visual cues, based on a trajectory generated using the method discussed in Section \ref{section:Gaze}, directly into the driver’s forward field of view, the HUD shows cues to guide their gaze and attention. Fig. \ref{fig:HUD} is an example of an HUD. 

Depending on the urgency and context of the hazard, the HUD could adapt its presentation style following the proposed visual cue hierarchy below:
\begin{itemize}
    \item \textbf{Pulsing Indicators}: Gentle blinking or pulsation for medium-priority warnings, such as a car moving into a blind spot, to emphasize importance without overwhelming the driver.
    \item \textbf{Color-Coded Overlays}: Gradual changes in the intensity of the color (e.g. yellow for caution or red for imminent threat) that reflect the severity of the hazard.
    \item \textbf{Animated Arrows or Icons}: Dynamic symbols point toward areas the driver may be overlooking, helping to break target fixation. 

\end{itemize}


For example, if the autonomous vehicle detects an unlabeled hazard outside the driver’s current focal area, the HUD projects sequential arrows or icons directing the driver's gaze and attention to critical areas of the scene. However, if the hazard escalates, the arrows or icons could change colors or begin pulsing to convey urgency to the driver. 

Various HUD technologies can deliver these alerts. For our proposal, a micro-LED or in-glass laminated HUD is best suited due to its high brightness, precise color control, and unobtrusive design—qualities essential for conveying urgent, context-rich information \cite{hud}.


 \subsection{Audio Alerts}
In addition to visual HUD cues, we propose employing strategically timed audio alerts to reinforce attention shifts toward potential hazards \cite{hidaka2015sound}. By leveraging the capacity of sound to capture immediate awareness

We leverage the capacity of sound to capture immediate awareness to further direct driver gaze, utilizing a potential audio alert system as outlined below.

\begin{itemize}
    \item \textbf{Low-Level Tones:} 
    Gentle, short-duration sounds serve as an early warning for a developing or moderate hazard. 
    \item \textbf{Urgent Beeps:} More pronounced and localized audio signals for high-priority hazards. 

\end{itemize}

This proposed audio alert scheme could mitigate target fixation and improve situational awareness, ensuring that even rapidly emerging hazards elicit swift recognition and response.

\section{Conclusion}
We aimed to improve driver readiness for scenarios containing out-of-label hazards by supporting existing human-vehicle interaction capabilities with new systems. This promotes the ability of an autonomous vehicle to prevent accidents during uncertain conditions, such as previously unseen hazards. In this paper, we propose a system using visual cues, shown in the driver's field of view via an HUD device, to shift driver attention to the unlabeled hazard and relevant areas in the scene. We use a saliency map, filtered to include relevant objects in relation to an unlabeled hazard, to generate a trajectory in which visual cues can be displayed for the driver to follow.
We believe this concept offers a promising avenue for improving situational awareness, mitigating risks from target fixation, and ensuring drivers are more responsive to emerging hazards. 
Future work includes developing a functional prototype and conducting comprehensive user studies in a high-fidelity simulation environment to validate the system's effectiveness in handling out-of-label hazards in autonomous driving.

\addtolength{\textheight}{-12cm}   




\bibliographystyle{ieeetr}
\bibliography{root}

\end{document}